\documentclass{article}

\usepackage[final]{corl_2022}

\usepackage{amsmath,amssymb}
\usepackage{dsfont}
\usepackage{array, booktabs}
\usepackage{multirow}
\usepackage{bbm}
\usepackage{graphicx}
\usepackage{tablefootnote}
\usepackage{subcaption}

\graphicspath{ {./figs/} }

\usepackage{amsmath,amsfonts,bm}

\def\eqref#1{equation~\ref{#1}}

\def\1{\bm{1}}

\DeclareMathAlphabet{\mathsfit}{\encodingdefault}{\sfdefault}{m}{sl}
\SetMathAlphabet{\mathsfit}{bold}{\encodingdefault}{\sfdefault}{bx}{n}

\newcommand{\E}{\mathbb{E}}

\DeclareMathOperator*{\argmax}{arg\,max}

\title{Hypernetworks in Meta-Reinforcement Learning}

\author{
  Jacob Beck\\
  Department of Computer Science\\
  University of Oxford,
  United Kingdom\\
  \texttt{jacob\_beck@alumni.brown.edu} \\
  \And
  Matthew Jackson \\
  Department of Engineering Science \\
  University of Oxford,
  United Kingdom \\
  \texttt{jackson@robots.ox.ac.uk} \\
  \And
  Risto Vuorio \\
  Department of Computer Science \\
  University of Oxford,
  United Kingdom\\
  \texttt{risto.vuorio@keble.ox.ac.uk} \\
  \And
  Shimon Whiteson \\
  Department of Computer Science \\
  University of Oxford,
  United Kingdom\\
  \texttt{shimon.whiteson@cs.ox.ac.uk} \\
}

\begin{document}
\maketitle











\begin{abstract}
    Training a reinforcement learning (RL) agent on a real-world robotics task remains generally impractical due to sample inefficiency.
    Multi-task RL and meta-RL aim to improve sample efficiency by generalizing over a distribution of related tasks.
    However, doing so is difficult in practice: In multi-task RL, state of the art methods often fail to outperform a degenerate solution that simply learns each task separately.
    Hypernetworks are a promising path forward since they replicate the separate policies of the degenerate solution while also allowing for generalization across tasks, and are applicable to meta-RL.
    However, evidence from supervised learning suggests hypernetwork performance is highly sensitive to the initialization.
    In this paper, we 1) show that hypernetwork initialization is also a critical factor in meta-RL, and that naive initializations yield poor performance; 2) propose a novel hypernetwork initialization scheme that matches or exceeds the performance of a state-of-the-art approach proposed for supervised settings, as well as being simpler and more general; and 3) use this method to show that hypernetworks can improve performance in meta-RL by evaluating on multiple simulated robotics benchmarks.

\end{abstract}



\section{Introduction}


Deep reinforcement learning (RL) has helped solve previously intractable problems but still remains highly sample inefficient.
This sample inefficiency makes it impractical, particularly in settings where data collection happens in the real world. 
For example, a robot's actions have the potential to inflict damage on both itself and its surroundings.
Multi-task RL and meta-RL aim to improve sample efficiency on novel tasks by generalizing over a distribution of related tasks. 
However, such generalization has proven difficult in practice.
In fact, multi-task RL methods often fail to outperform a degenerate solution that simply trains a separate policy for each task \citep{meta-world}.

One promising way to improve generalization is with a \textit{hypernetwork}, a neural network that produces the parameters for another network, called the \textit{base network} \citep{ha2017hypernetworks}.
In multi-task RL, using a hypernetwork that conditions on the task ID to generate task-specific parameters can replicate the separate policies of the degenerate solution, while also allowing generalization across tasks.
Furthermore, unlike the degenerate solution, hypernetworks can also be applied to meta-RL, where task IDs are not provided and test tasks may be novel, by conditioning them on the output of a task encoder.

However, hypernetworks come with their own challenges.
Since hypernetworks generate base network parameters, the initialization of parameters in the hypernetwork determines the initialization of the base network it produces.
Evidence suggests hypernetwork performance is highly sensitive to the initialization scheme in supervised learning \citep{hfi}. However, to our knowledge this question has not been considered in meta-RL. In this paper, we show that hypernetwork initialization is also a critical factor in meta-RL, and that naive initializations yield poor performance.


\begin{figure}[t]
    \centering
    \begin{subfigure}[t]{0.5\textwidth}
        \centering
        \includegraphics[width=\linewidth]{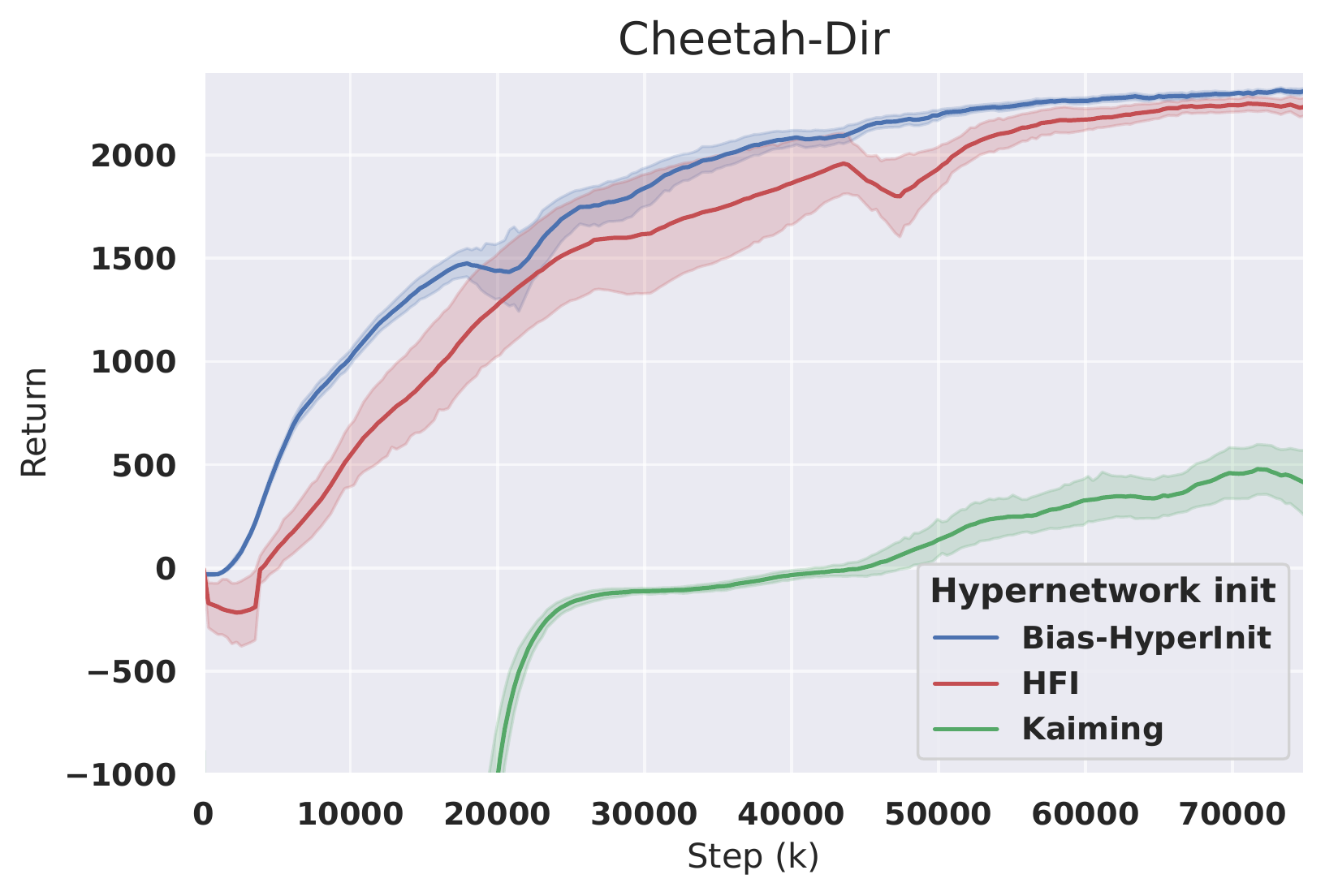}
        \caption{Hypernetwork initialization methods}
    \end{subfigure}%
    \begin{subfigure}[t]{0.475\textwidth}
        \centering
        \includegraphics[width=\linewidth]{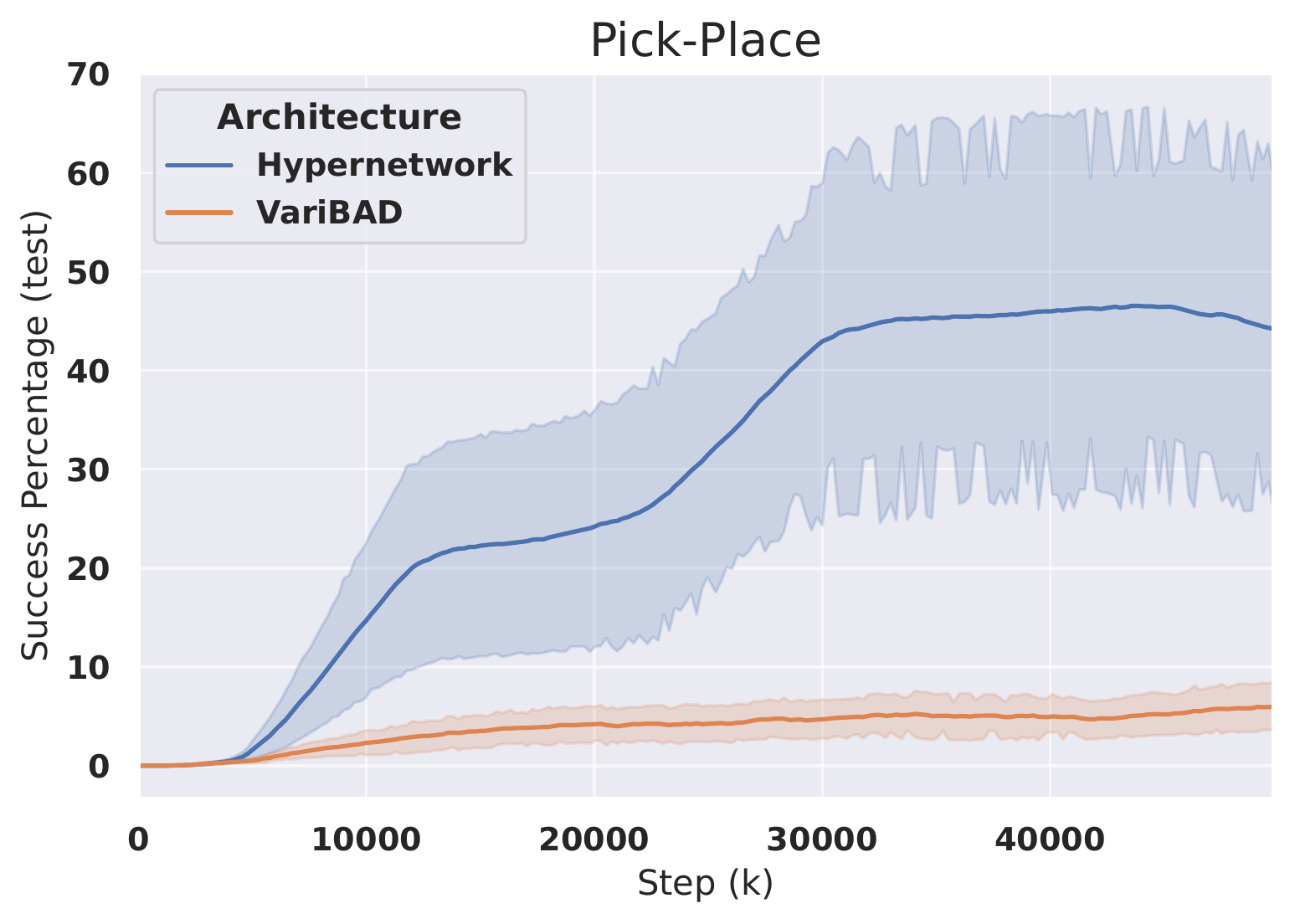}
        \caption{Architectures}
    \end{subfigure}
    \caption{Naive initializations such as Kaiming \citep{kaiming} fail for hypernetworks, whereas our proposed Bias-HyperInit does not and matches the state of the art, HFI \citep{hfi} (claims 1, 2). Adding hypernetworks with the proposed Bias-HyperInit significantly improves the state-of-the-art meta-RL method, VariBAD \citep{varibad} (claim 3).}
    \label{fig:lr_curves}
\end{figure}

Furthermore, we propose two novel initialization schemes: \textit{Bias-HyperInit} and \textit{Weight-HyperInit}.
Both produce strong results, with the former matching or exceeding the performance of the state-of-the-art hypernetwork initialization method designed for supervised learning \citep{hfi}.
Moreover, both proposed methods are simpler and more general than this existing method, in that they may be applied to arbitrary base network architectures and target base network initializations without additional derivation.
Using Bias-HyperInit, we present results that substantially improve the a state-of-the-art method on a range of meta-RL benchmarks.

Applying hypernetworks to meta-RL, we make the following contributions (examples in \autoref{fig:lr_curves}):
\begin{enumerate}
    \item We empirically demonstrate that initialization is a critical factor in the performance of hypernetworks in meta-RL, and that naive initializations fail to learn reliably;
    \item We propose a novel hypernetwork initialization scheme that matches or exceeds the performance of a state-of-the-art approach proposed for supervised settings, as well as being simpler and more general; and
    \item We use this method to show that hypernetworks can improve a state-of-the-art method on a range of meta-RL benchmarks (grid-world \citep{varibad}, MuJoCo \citep{mujoco}, and Meta-World \citep{meta-world}).
\end{enumerate}



\section{Related Work}
\label{sec:related_work}


\paragraph{Meta-RL.} Despite the advantages of hypernetworks \citep{ha2017hypernetworks}, they remain relatively unexplored in meta-RL. 
We use hypernetworks to arbitrarily update a policy's parameters at every time-step, whereas all prior work we are aware of restrict this procedure in some way.
Many procedures in few-shot meta-RL build off of MAML \citep{maml} to adapt the parameters of a policy network using a policy gradient \citep{maml, mmaml, metaSGD}.
Such methods require the estimation of a policy gradient, which reduces sample-efficiency when faster adaptation is possible, as in our benchmarks \citep{varibad}.
Most meta-learning procedures capable of zero-shot adaption using an RNN (or convolutions) that can represent an arbitrary update function \citep{varibad, snail, rl2}.
These methods generally update a set of activations on which a fixed policy is then conditioned, whereas hypernetworks update all policy parameters.
We include a state-of-the-art method from this class in our evaluations \citep{varibad}.
There are also unsupervised methods in zero-shot meta-RL for weight updates \citep{hebb, diff_plastic} but none can produce a fully general learning procedure since they make use of local and unsupervised heuristics. \citet{sarafian2021recomposing} use hypernetworks in the context of meta-RL, but the policy network, not the hypernetwork, is conditioned on the RNN used for adaptation, preventing the hypernetwork from representing a general learning procedure.
Finally, FLAP \citep{flap} learns to infer a set of weights trained in the multi-task setting; however since the adaptation procedure is not trained on a meta-RL objective, it is constrained. For example, FLAP cannot learn to explore to reduce uncertainty.
Finally, \citet{xian2021hyperdynamics} use hypernetworks to predict model dynamics then use model predictive control. However, this model still requires planning to make use of an uncertain model, whereas model-free RL learns a policy that explores optimally in order to attain data for adaptation.
Using a general procedure trained to arbitrarily modify the weights of a model-free policy has never been tried in RL, to the best of our knowledge.

\paragraph{Hypernetworks.} 
Hypernetworks, or similar architectures, have been used in supervised learning (SL), multi-task RL, and meta-SL.
Hypernetworks have been used in the supervised learning literature for sequence modelling \citep{ha2017hypernetworks}, as well as in continual learning and image classification \citep{hfi}, where it was shown that the hypernetwork initialization scheme was crucial for performance.
Similar models have also been used in multi-task RL and meta-SL, but not meta-RL.
For instance, in multi-task RL, \citet{no_interfere} use a network conditioned on a task encoding to produce the weights and biases for every other layer in another network conditioned on state.
In meta-SL, there have also been attempts to use one network to adapt weights of another, both as a general function of the dataset \citep{leo, metanet, hypermaml}, conditioned on an embedding adapted by gradient descent \citep{meta_via_hyper}, and by adding deltas in a way framed as learning to optimize \citep{ravi2017optimization, li2017learning}.
The abundance of representations in meta-SL suggest there is a similarly large space of representation-based methods to explore in meta-RL.
Our work -- getting hypernetworks to work in practice for meta-RL -- can be seen as a first step towards applying all of these methods in meta-RL.


\section{Background}


\subsection{Problem Setting}
An RL task is formalized as a Markov Decision Processes (MDP). We define an MDP as a tuple of $(\mathcal{S, A, R, P}, \gamma)$. At time-step $t$, the agent inhabits a state, $s_t \in \mathcal{S}$, observable by the agent. The agent takes an action $a_t \in \mathcal{A}$. The MDP then transitions to state $s_{t+1} \sim \mathcal{P}(s_{t+1} | s_t, a_t)\colon \mathcal{S} \times \mathcal{A} \times\mathcal{S} \rightarrow \mathds{R}_{\ge 0}$, and the agent receives reward $r_{t} = \mathcal{R}(s_t, a_t)\colon \mathcal{S} \times \mathcal{A} \rightarrow \mathds{R}$ upon entering $s_{t+1}$. Given a discount factor, $\gamma \in [0, 1)$, the agent acts to maximize the expected future discounted reward, $R(\tau) = \sum_{r_t \in \tau}{\gamma^t r_t}$, where $\tau$ is the agent's trajectory over an episode in the MDP.
To maximize this return, the action takes actions sampled from a learned policy, $\pi(a|s): \mathcal{S} \times \mathcal{A} \rightarrow \mathbb{R}_+.$

Meta-RL algorithms learn an RL algorithm, i.e., a mapping from the data sampled from a single MDP, $\mathcal{M} \sim p(\mathcal{M})$, to a policy. Since an RL algorithm generally needs multiple episodes of interaction to produce a reasonable  policy, the algorithm conditions on $\tau$, which is the entire sequence of states, actions, and rewards within $\mathcal{M}$. As in the RL setting, this sequence up to time-step t forms a trajectory $\tau_t \in (\mathcal{S} \times \mathcal{A} \times \mathbb{R})^{t} $.
Here, however, $\tau$ may span multiple episodes, and so we use the same symbol, but refer to it as a \textit{meta-episode}.  
The policy is then a meta-episode dependent policy, $\pi_\theta(a|s,\tau)$, parameterized by the \textit{meta-parameters}, $\theta$.

We define the objective in meta-RL as finding meta-parameters $\theta$ that maximize the sum of the returns in the meta-episode across a distribution of tasks (MDPs):
\begin{align}\label{eq:objective}
    \argmax_{\theta} \E_{\mathcal{M} \sim p(\mathcal{M})} \bigg[ \E_\tau \bigg[ R(\tau) \bigg| \pi_\theta(\cdot), \mathcal{M} \bigg]\bigg]
\end{align}



\subsection{Policy Architecture}
\label{sec:background_arch} 

We consider meta-RL agents capable of adaptation at every time-step, and adaptation within one episode is required to solve some of our benchmarks.
In such methods \citep{varibad, snail, rl2}, the history is generally summarized by a function, $g$, into an embedding that represents relevant task information. 
We write this embedding as $e=g(\tau)$, and call $g$ the task encoder.
The policy, represented as a multi-layer perceptron, then conditions on this task embedding as an input, instead of on the history directly, which we write as $\pi_\theta(a|s,e)$.
We call this the \textit{standard architecture}, shown in \autoref{fig:arch}.

In this paper, we primarily build off of VariBAD \citep{varibad}, which can be seen as an instance of the standard architecture where the task encoder is the mean and variance from a recurrent variational auto-encoder (VAE) \cite{vae} trained using a self-supervised loss.
In other words, the task is inferred as a latent variable optimized for reconstructing a meta-episode.
See \citet{varibad} for details.
Additionally, evaluate the addition of hypernetworks to RL2 \citep{rl2} on the most challenging benchmark. 
(See section \ref{sec:metaworld}.)
In RL2, the task encoder is a recurrent neural network trained end-to-end on equation \ref{eq:objective}. 

\subsection{Hypernetwork Initialization}

\citet{hfi} show that applying existing initialization methods for neural networks to hypernetworks produces unstable base network initializations with exploding or vanishing activations.
Furthermore, they empirically demonstrate a reduction in training stability for Kaiming initialization \citep{kaiming}.
We corroborate this failure for Kaiming initialization on meta-RL, as well as for Orthogonal initialization \citep{orthogonal} and Normc initialization \citep{baselines}, which we collectively refer to as \textit{default initializations.}

As a solution, \citet{hfi} propose the first initialization designed for hypernetworks and show it to be effective in supervised learning.
Their approach is based on Kaiming initialization, which samples network parameters such that the activations of the network at each layer maintain the same variance as in the previous layer.
\citet{hfi} extend this variance analysis to hypernetworks \citep{ha2017hypernetworks}. 
They propose two methods: Hyperfan-in (HFI) sets the variance of the initial parameter distribution of the hypernetwork to maintain constant variance of activations in the base network in the forward pass, and hyperfan-out (HFO) does the same for the backward pass.
Since these are equally competitive, and produce state-of-the-art results, we include HFI as a baseline for comparison.
However, this variance analysis is involved and requires modification for specific use cases depending on the activation function and whether or not the network produces weights or biases in the base network. 
This motivates the need for a simpler and more general initialization method, which we propose in this paper.

\section{Methods}

In this section we introduce our proposed architecture using hypernetworks and our proposed hypernetwork initialization. At a high level, the hypernetwork conditions on a task representation to generate all of the parameters for a base policy. The initialization provides a simple and general way to sample parameters for this hypernetwork at the start of training. 

\subsection{Policy Architecture}

\begin{figure}[t]
\begin{center}
\includegraphics[width=\linewidth]{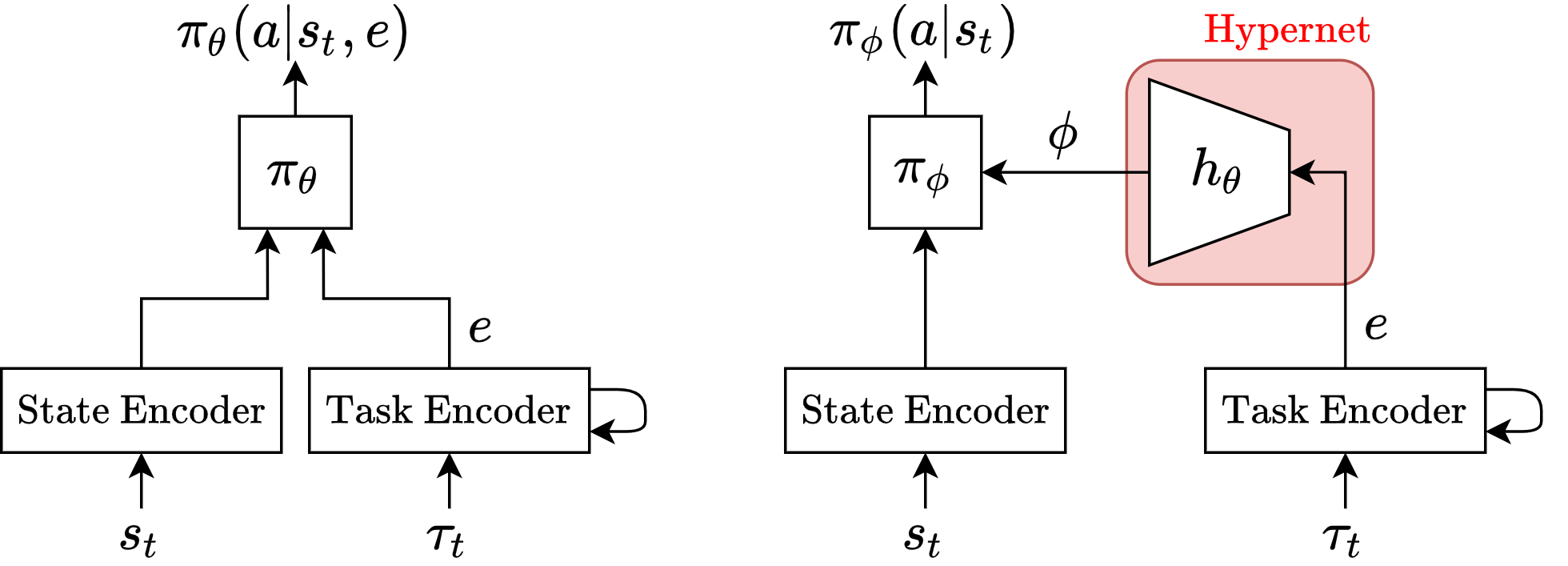}
\end{center}
\caption{A standard architecture (left) and hypernetwork model (right).}
\label{fig:arch}
\end{figure}

We propose the use of hypernetworks in meta-RL, instead of the standard architecture described earlier. 
In this setting, we use a hypernetwork to arbitrarily adapt the parameters of the base policy.
We still use a task encoder, $e=g(\tau)$, but instead of conditioning a policy on these activations, we use a hypernetwork, $h_\theta$, to generate policy parameters, $\phi=h_\theta(e)$.
These parameters, i.e., weights and biases, are then used directly for the base network, which we write: $\pi_\phi(a|s)$.
This is also depicted in \autoref{fig:arch}.

In this case, the hypernetwork can arbitrarily adapt all the the parameters of $\pi$ based on history.
In comparison, in the standard architecture, the shared fixed parameters of the base network ($\theta)$ are still required to generalize between all of the tasks. 
Since training separate policies for each task often performs better than state-of-the-art methods for generalizing across all tasks \citep{meta-world}, this motivates the ability to produce base policies with no or few shared parameters.
Hypernetworks allow for shared parameters when possible, but also provide the ability to have no shared parameters in the base policy when diverse policies are necessary and little generalization between tasks is possible.


In fact, hypernetworks are capable of replicating the training of separate policies for each task under certain conditions.
To see this, consider the case where the hypernetwork is linear and has no bias.
Then, the hypernetwork consists only of a weight matrix, $W$. (That is, $\theta = W$.)
If this hypernetwork conditions on a one-hot task ID for task i: $e=\mathbbm{1}_i$, then
 the parameters selected by this hypernetwork, $\phi$ are a separate set of parameters for each task: $\phi^i = h(e) = W \mathbbm{1}_i$.
In other words, training individual networks for each task is equivalent to training a hypernetwork when that hypernetwork is: 1) linear, 2) has no a bias, and 3) is conditioned on a one-hot task ID. 

However, we can relax these assumptions and still retain the ability to produce distinct parameters, while also enabling generalization.
If we add a bias, we reintroduce shared parameters in the hypernetwork, but they can still produce separate base networks for each task when little transfer between the policies is required. 
If we relax the one-hot assumption, the network is still capable of producing a one-hot encoding when the tasks are discrete and the task embedding is sufficiently large.
Relaxing these restrictions allows for both generalization and the application of hypernetworks to meta-RL.

\subsection{Hypernetwork Initialization}

Default initialization methods fail for hypernetworks. However, given that hypernetworks generalize training separate networks for each task, it must be possible to initialize them as if each of the corresponding base networks were initialized independently, from some given initialization function, $f$, known to train reliably.
Using this insight, we propose and evaluate two initialization schemes for the hypernetwork. We propose one where (under some assumptions) each base network is independently initialized from $f$. We also propose one where (under no assumptions) all base networks share an initialization sampled from $f$. 

Our first method is \emph{Weight-HyperInit}. Let $W$ and $b$ be the weights and bias in the last layer of our hypernetwork, $h(e)$, respectively. (These parameters are both contained in $\theta$.) We define this weight-only initialization as follows:
$$W_{:,i} := \phi^i \sim f(\phi) \:\: \forall i,\qquad b := \mathbf{0}, $$
where $f$ is an arbitrary initialization scheme for the base network specifying a distribution over a vector of parameters, $\phi$, and $W_{:, i}$ is the $i$-th column of $W$. 

Weight-HyperInit reproduces any given initialization for each base network ($\pi_\phi$), under the assumptions that $e=\mathbbm{1}_i$ and the hypernetwork is linear.
In this case, each column of $W$ is simply a sample from the base scheme, one of which is selected for each task via the one-hot encoding. For example, given the task embedding $e = \mathbbm{1}_3$, the following base network initialization $\phi_{\text{init}}$ is produced:
\begin{equation}
    \phi_{\text{init}} = We+b =
    \begin{pmatrix}
        \phi^1_1 & \phi^2_1 & \phi^3_1 & \dots \\ 
        \phi^1_2 & \phi^2_2 & \phi^3_2 & \dots \\ 
        \phi^1_3 & \phi^2_3 & \phi^3_3 & \dots \\ 
        \vdots & \vdots & \vdots & \ddots
    \end{pmatrix}
    \begin{pmatrix}
        0 \\ 0 \\ 1 \\ 0 \\ \vdots
    \end{pmatrix}
    + \mathbf{0} = \phi^3.
\end{equation}
Moreover, in the case that there is also no bias, it is also equivalent to training separate networks for each task. Although these assumptions are not met for meta-RL, and so do not hold for our experiments, we find this is still an improvement over default neural network initialization schemes.


Additionally, we propose \emph{Bias-HyperInit}. This bias-only initialization is defined as follows:
$$ W_{i,j} := 0 \:\: \forall i,j, \qquad b := \phi_{\text{shared}} \sim f(\phi).$$
Bias-HyperInit achieves an arbitrary initialization for the base network without any assumptions, by setting the parameters for any task to be the same at initialization.
This encourages parameter sharing between base networks at the beginning of training, where possible, but also allows for separate base network parameters to be learned, if necessary.
Under any set of assumptions, the base network is initialised to the following:
\begin{equation}
    \phi_{\text{init}} = Wx+b =
    \begin{pmatrix}
        0 & 0 & 0 & \dots \\ 
        0 & 0 & 0 & \dots \\ 
        0 & 0 & 0 & \dots \\ 
        \vdots & \vdots & \vdots & \ddots
    \end{pmatrix}
    \begin{pmatrix}
        x_1 \\ x_2 \\ x_3 \\ x_4 \\ \vdots
    \end{pmatrix}
    + \phi_{\text{shared}} = \phi_{\text{shared}},
\end{equation}
where $x$ is the final hidden layer of the hypernetwork. ($X=e$ in the case of a linear hypernetwork.)

Both methods initialize only $W$ and $b$, which define the head of the network. All other layers in $h$ may be initialized by any default initialization scheme.
All such choices are detailed in supplementary materials.

\subsection{Baselines}

\paragraph{VariBAD \& RL2.} See Sec.\ \ref{sec:background_arch}.


\paragraph{FiLM.} FiLM \citep{film} is a convenient baseline situated between hypernetworks and the standard architecture. In FiLM, the hypernetwork generates biases for each layer, but only point-wise scales the activations instead of generating weight matrices.
In this case, the base network has its own weights.
Bias-HyperInit can easily be adapted to FiLM; details presented in supplementary materials.


\paragraph{HFI.} 
HFI \citep{hfi} is a state-of-the-art initialization method developed tested in the supervised learning setting.
While we do compare to HFI, our methods are simpler and more general. 
Specifically, our methods work with arbitrary base network target initialization (as opposed to being tied to Kaiming) and our methods work with arbitrary base network architectures (without additional variance analysis). 
Moreover, our approach is straightforward to apply to additional methods, which we show by applying it to FiLM. Finally, although both HFI and Weight-HyperInit do make assumptions about the input to the hypernetwork, our strongest method, Bias-HyperInit, does not.



\section{Experiments}
\label{sec:result}





In this section, we compare our hypernetwork architecture and initialization methods to baselines on 2D navigation \citep{varibad}, MuJoCo \citep{meta-world}, ML1 \citep{meta-world}, and ML10 \citep{meta-world} benchmarks.
MuJoCo is a common meta-RL benchmark \citep{maml, varibad, humplik2019meta, rakelly2019efficient, hebb}, as are toy 2D navigation tasks \citep{maml, varibad, humplik2019meta, rakelly2019efficient}. 
These two benchmarks allow us to demonstrate that hypernetworks with default initialization methods fail to learn, whereas our proposed methods learn reliably.
ML1 and ML10 are benchmarks in Meta-World \citep{meta-world}.
These two benchmarks have greater room for improvement with state-of-the-art methods \citep{varibad}, which allow us to demonstrate improvement over the baseline architectures.
Finally, we use two MuJoCo environments to investigate the performance of hypernetworks against standard architectures in terms of the number of parameters in the model overall.

Throughout our evaluation, we use two-tailed $t$-tests with $p=0.05$ to determine significance. Details on model tuning and implementation are presented in supplementary materials.

\subsection{Navigation and MuJoCo}
Here we evaluate on the grid-world variant from \citet{varibad} as our 2D navigation task and MuJoCo \citep{mujoco}. Note Grid-World and Cheetah-Dir contain twenty-four and two non-parametric tasks respectively, while the other MuJoCo environments have parametric variation between the tasks.

In \autoref{tab:grid_mujoco} we see that default initializations frequently fail to learn while Bias-HyperInit learns reliably.
Specifically, Kaiming and Normc initializations used with hypernetworks achieve far lower returns than all other methods.
Orthogonal initialization is more competitive, however it is still significantly outperformed by Bias-HyperInit in every environment.

We also compare hypernetworks with Bias-HyperInit to the standard architecture and Bias-HyperInit to HFI.
We see that Bias-HyperInit matches or exceeds the performance of HFI, with a significant improvement in Walker.
Hypernetworks with Bias-HyperInit also significantly outperform the standard architecture on grid-world and Ant-Dir.
In fact, Bias-HyperInit is not significantly outperformed by any other method.
However, the standard architecture, HFI, and Bias-HyperInit all achieve near optimal performance, motivating an evaluation on Meta-World, on which the standard architecture has greater room for improvement \citep{varibad}.

\newcolumntype{R}{>{\raggedleft}p{2.5em}}
\begin{table}[t!]
    \centering
    \caption{Comparison of return on grid-world and MuJoCo tasks over five random seeds (mean $\pm$ standard error). Entries in bold have insignificant difference from the highest-performing result.}
    \begin{tabular}{@{} l*{5}{R@{\hskip 0.7mm}c@{\hskip 0.7mm}r} @{}}
        \toprule
        \textbf{Method} & \multicolumn{3}{r}{\textbf{Grid-World}} & \multicolumn{3}{r}{\textbf{Cheetah-Dir}} & \multicolumn{3}{r}{\textbf{Walker}} & \multicolumn{3}{r}{\textbf{Ant-Dir}} & \multicolumn{3}{r@{}}{\textbf{Humanoid}} \\
        \midrule
        Standard & $35.5 $&$\pm$&$ 0.4$ & $\mathbf{2104}$&$\pm$&$\mathbf{87}$ & $\mathbf{1828} $&$\pm$&$\mathbf{38}$ & $1167 $&$\pm$&$ 16$ & $\mathbf{1842}$&$\pm$&$\mathbf{233}$ \\
        \cmidrule{2-16}
        Kaiming & $32.1$&$\pm$&$1.2$ & $378$&$\pm$&$169$ & $331$&$\pm$&$37$ & $253$&$\pm$&$86$ & $266$&$\pm$&$23$ \\
        Normc & $32.2$&$\pm$&$0.5$ & $356$&$\pm$&$134$ & $357$&$\pm$&$60$ & $264$&$\pm$&$106$ & $249$&$\pm$&$18$ \\
        Orthogonal & $34.9$&$\pm$&$0.5$ & $1379$&$\pm$&$310$ & $1687$&$\pm$&$98$ & $1127$&$\pm$&$93$ & $1126$&$\pm$&$76$ \\
        HFI & $\mathbf{36.8}$&$\pm$&$\mathbf{0.2}$ & $\mathbf{2218}$&$\pm$&$\mathbf{80}$ & $1618$&$\pm$&$130$ & $\mathbf{1370}$&$\pm$&$\mathbf{9}$ & $\mathbf{1323}$&$\pm$&$\mathbf{57}$ \\
        \cmidrule{2-16}
        Weight-HyperInit & $\mathbf{36.1}$&$\pm$&$\mathbf{0.5}$ & $\mathbf{2066}$&$\pm$&$\mathbf{119}$ & $1748$&$\pm$&$57$ & $\mathbf{1346}$&$\pm$&$\mathbf{7}$ & $1048$&$\pm$&$21$ \\
        Bias-HyperInit & $\mathbf{36.7}$&$\pm$&$\mathbf{0.2}$ & $\mathbf{2300}$&$\pm$&$\mathbf{32}$ & $\mathbf{1994}$&$\pm$&$\mathbf{67}$ & $\mathbf{1328}$&$\pm$&$\mathbf{23}$ & $\mathbf{1678}$&$\pm$&$\mathbf{162}$ \\
        \bottomrule
    \end{tabular}
    \label{tab:grid_mujoco}
\end{table}

\vspace{.2cm}
\subsection{ML1 and ML10}
\label{sec:metaworld}
\vspace{.2cm}


Here we evaluate on the more challenging Meta-World ML1 and ML10 benchmarks \citep{meta-world}. 
ML1 and ML10 have one and ten non-parametric training tasks respectively (e.g. pushing a ball or opening a window).
ML10 additionally has five distinct test tasks. 
Within each task, there exists parametric variation, such variation in the goal location.
Note that we only test on the Pick-Place task from ML1, since VariBAD already achieves a $100\%$ success rate on all other tasks \citep{varibad}.


\begin{table}[ht]
    \centering
    \caption{Comparison of meta-test success percentage on the Pick-Place ML1 task (ten seeds) and ML10 (three seeds).}
    \begin{tabular}{@{} ll*{3}{R@{\hskip 0.7mm}c@{\hskip 0.7mm}r} @{}}
        \toprule
        \multirow{2}{*}{\textbf{Method}} & & \multicolumn{3}{c}{\textbf{Pick-Place}} & \multicolumn{6}{c@{}}{\textbf{ML10}} \rule[-2mm]{0pt}{0pt} \\
        & & \multicolumn{3}{c}{\textbf{VariBAD}} & \multicolumn{3}{c}{\textbf{VariBAD}} & \multicolumn{3}{c@{}}{\textbf{RL2}} \\
        \midrule
        Standard & & $4.4 $&$\pm$&$ 2.4$\tablefootnote{\citet{varibad} report a success percentage of 29\% for Pick-Place; however, we were not able to replicate this result. 
        } & $10.2 $&$\pm$&$ 3.0$ & $7.2 $&$\pm$&$ 5.0$ \\
        \cmidrule{2-11}
        \multirow{2}{*}{FiLM} & Normc & $5.5$&$\pm$&$4.8\:\:$ & &---& & &---& \\
        & Bias-HyperInit & $\mathbf{34.2}$&$\pm$&$\mathbf{15.9}\:\:$ & &---& & &---& \\
        \cmidrule{2-11}
        \multirow{2}{*}{Hypernetwork} & HFI & $\mathbf{25.5}$&$\pm$&$\mathbf{14.5}\:\:$ & $28.4 $&$\pm$&$6.0$ & $7.1 $&$\pm$&$2.4$ \\
        & Bias-HyperInit & $\mathbf{42.9 }$&$\pm$&$\mathbf{ 16.3}\:\:$ & $23.9 $&$\pm$&$ 6.2$ & $14.2 $&$\pm$&$ 7.2$ \\
        \bottomrule
    \end{tabular}
    \label{tab:meta_world}
\end{table}
\vspace{.4cm}

In \autoref{tab:meta_world} we see significant improvement from hypernetworks with Bias-HyperInit over the standard architecture, as well as the efficacy of Bias-HyperInit on FiLM.
On Pick-Place, Bias-HyperInit outperforms the standard architecture with a 9-fold increase in test success percentage.
Additionally, Bias-HyperInit improves the FiLM architecture and exceeds the performance of HFI.
On ML10, hypernetworks with both Bias-HyperInit and HFI yield a 2-fold increase in test success percentage compared to the standard architecture.
Finally, we evaluated Bias-HyperInit when applied to RL2 on ML10, finding a 2-fold increase over both HFI and the standard architecture.
Taken together, these results show a clear improvement from the application of hypernetworks with Bias-HyperInit, regardless of the baseline method they are applied to.

\newpage

\subsection{Network Size Comparison}


\begin{figure}
    \centering
    \includegraphics[width=\linewidth]{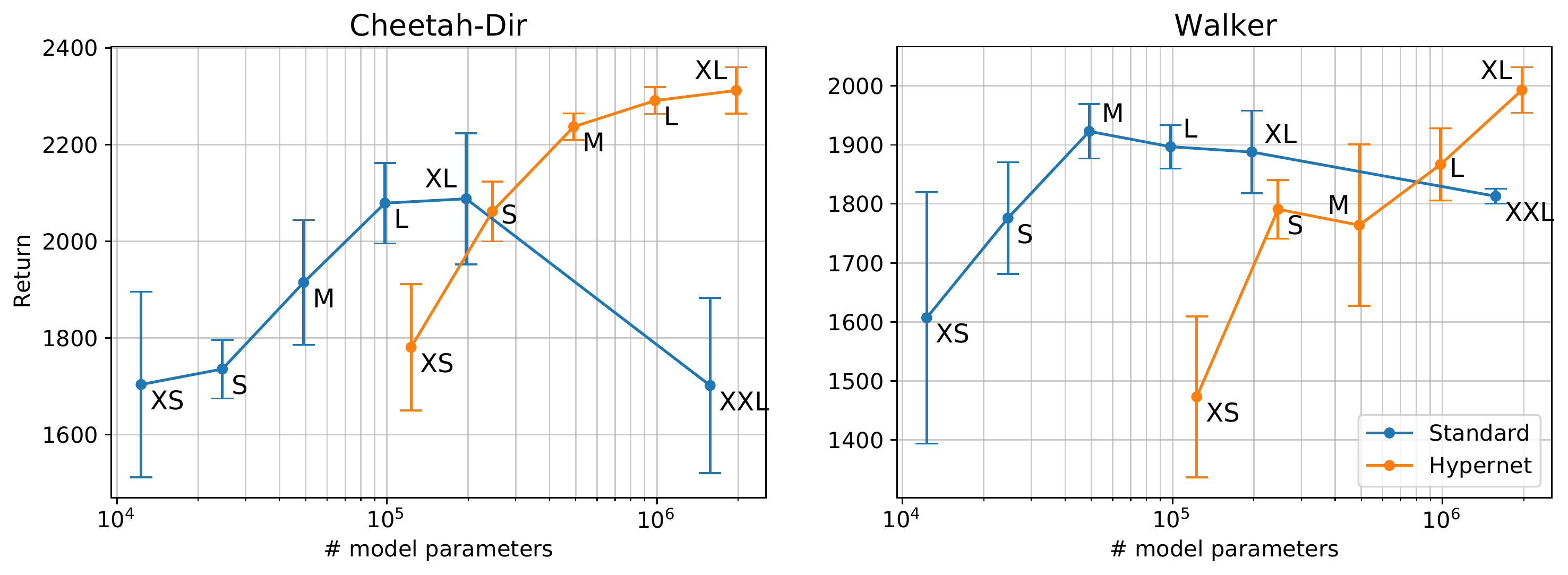}
    \caption{Performance of standard and hypernetwork models over a range of base policy architecture sizes on Cheetah-Dir and Walker. Architectures are presented in supplementary materials.}
    \label{fig:scaling}
\end{figure}

Because hypernetworks learn a mapping from a task embedding to the parameters of a base network,
they require significantly more parameters in the entire model than a standard architecture with the equivalent base network.
For a fair comparison, we evaluate return over both a range of base network sizes and total number of parameters in the model on the Cheetah-Dir and Walker tasks (\autoref{fig:scaling}).
We find that hypernetworks consistently equal or outperform the standard architectures with the equivalent base network size.
We also find that hypernetworks likewise outperform standard architectures for equivalent number of parameters in the entire model, i.e. for a given value on the x-axis, when the total number of parameters in the model is sufficiently large.

\section{Limitations}
\label{sec:limitations}
While our proposed methods are general, we cannot guarantee an improvement for all meta-RL methods.
To mitigate this limitation, we build on top of VariBAD, which is state of the art, and additionally evaluate our method applied to RL2 on ML10.
Furthermore, as in any empirical study, there is no guarantee that our results hold on real robots or other meta-RL benchmarks.
However, we have tested on seven standard meta-RL environments in total, including Meta-World, which was proposed specifically for addressing robotics.
As much as is possible from simulated meta-RL experiments, these results give us confidence in a significant improvement over previous methods.


\section{Conclusion}
\label{sec:conclusion}
We used hypernetworks to improve a state-of-the-art method in meta-RL, evaluating over a range of benchmarks. 
In doing so, we demonstrated that hypernetworks are a promising path forward for meta-RL. 
Moreover, we showed that the initialization of the hypernetwork is crucial, as default initialization methods fail.
To overcome this difficulty, we presented two novel initialization methods: Bias-HyperInit and Weight-HyperInit.
Bias-HyperInit matched or exceeded the performance of existing methods from the supervised learning setting, while also being simpler and more general -- applying to arbitrary base network initializations, base network architectures, and also improving FiLM.
Using Bias-HyperInit, we showed that hypernetwork performance improves substantially over standard architectures.
Finally, we demonstrated that hypernetworks outperform the standard architecture for equivalently sized base policies, and outperform it at any size given sufficiently many parameters in the entire model.
This paper additionally opens the path for future research extending meta-SL methods using hypernetworks \citep{leo, ravi2017optimization} and multi-task RL methods with separate parameters for each task \citep{distral,actor_mimic,cross} to meta-RL.

\clearpage
\acknowledgments{We would like to thank Luisa Zintgraf for her help with the VariBAD code-base along with general advice and discussion. Jacob Beck is supported by the Oxford-Google DeepMind Doctoral Scholarship. Matthew Jackson is supported by the UK EPSRC CDT in Autonomous Intelligent Machines and Systems, with funding from AWS in collaboration with the Oxford-Singapore HMC Initiative. Risto Vuorio is supported by EPSRC Doctoral Training Partnership Scholarship and Department of Computer Science Scholarship.}


\bibliography{refs}  

\clearpage

\end{document}


\maketitle









\section{Benchmark Details}
We evaluate on grid-world \citep{varibad}, MuJoCo \citep{mujoco}, and Meta-World \citep{meta-world}. For Meta-World\footnote{\url{https://github.com/rlworkgroup/metaworld}}, we use version two of ML10 and version one of Pick-Place (ML1).
We use gridworld and MuJoCo environments from the reference VariBAD implementation\footnote{\url{https://github.com/lmzintgraf/varibad}}.
The number of episodes per meta-episode are the same as in this implementation. Returns reported are summed across episodes in the meta-episode.

\section{Model Implementation}

\subsection{Network Architecture}
For all experiments, we use a three-layer MLP network to represent the base network, either learning the parameters directly or generating them with a hypernetwork. Both of these networks are conditioned on a task embedding of size 10 in all experiments. A range of architecture sizes were explored on Cheetah-Dir and Walker MuJoCo tasks, which are detailed in \autoref{tab:archs}. 

\begin{table}[ht]
    \centering
    \caption{Policy (base) network architectures.}
    \begin{tabular}{@{} lr@{\,}r@{\,}r @{}}
        \toprule
        \textbf{Size} & \multicolumn{3}{r @{}}{\textbf{Layer widths}} \\
        \midrule
        XS & (64, & 64, & 32) \\
        S & (128, & 64, & 64) \\
        M & (128, & 128, & 64) \\
        L & (256, & 128, & 128) \\
        XL (\textit{Default}) & (256, & 256, & 128) \\
        XXL & (1024, & 512, & 512) \\
        \bottomrule
    \end{tabular}
    \label{tab:archs}
\end{table}

\subsection{Actor and Critic Networks} In the standard architecture, we use separate actor and critic networks. In the hypernetwork model, there are separate heads on the hypernetwork for the actor and critic parameters. Since all reported results use linear hypernetworks, this is equivalent to using separate hypernetworks for the actor and critic, conditioned on the same task embedding.

\subsection{Choice of Base Network Initialization} 
Our proposed methods, Bias-HyperInit and Weight-HyperInit, require a given initialization method for the base network, which we denote $f$.
We chose normc intialization \cite{baselines} for $f$, as it is the default initialization in the reference VariBAD implementation.
Note that, as in VariBAD, the state and task encoders use Kaiming initialization with a uniform distribution \citep{kaiming} (or orthogonal initialization in the RNN), and the head of the critic (produced by the hypernetwork) also uses Kaiming.
Additionally, the head of the actor has a gain of 0.01 for categorical distributions (i.e.\ in gridworld), and 1 for continuous distributions (as is standard for linear layers), as is default in VariBAD. 
However, as this default was noticed after experiments, the actor and critic heads of both HyperInit methods have a gain of $\sqrt{2}$, which is the default for ReLU activations.

\section{Hyperparameter Tuning}
Hyperparameters are identical to those in the reference VariBAD implementation, other than the learning rate and network architecture. 

For Cheetah-Dir and Walker, the models were tuned over the following set of learning rates on three seeds: $\{3\mathrm{e}{-3}, 1\mathrm{e}{-3}, 3\mathrm{e}{-4}, 1\mathrm{e}{-4}, 3\mathrm{e}{-5}\}$.
Learning rates of $1\mathrm{e}{-3}$ and $1\mathrm{e}{-4}$ yielded the best performance for standard and hypernetwork models respectively, over both domains.
For the remainder of the MuJoCo tasks and grid-world, the learning rate was tuned over a narrower range of learning rates.
These ranges were $\{3\mathrm{e}{-3}, 1\mathrm{e}{-3}, 3\mathrm{e}{-4}\}$ for the standard model and $\{3\mathrm{e}{-4}, 1\mathrm{e}{-4}, 3\mathrm{e}{-5}\}$ for the hypernetwork.
Again, we found that $1\mathrm{e}{-3}$ and $1\mathrm{e}{-4}$ yielded the best performance across domains for the standard and hypernetwork models respectively.
Given this, we used these two tuned learning rates for further evaluation on Pick-Place.
On ML10, we again evaluate over the full range: $\{3\mathrm{e}{-3}, 1\mathrm{e}{-3}, 3\mathrm{e}{-4}, 1\mathrm{e}{-4}, 3\mathrm{e}{-5}\}$.

The network sizes were tuned on Cheetah-Dir and Walker as well.
The standard and hypernetwork models were turned over all base network sizes in \autoref{tab:archs}, with the exception of the XXL base model, which was included only to demonstrate the performance of the standard model with an equal number of model parameters as the XL hypernetwork architecture.
For both hypernetworks and the standard network, the XL size achieved the highest return (or not significantly different from the best) on both tasks.
We then used XL for the rest of the MuJoCo tasks and for Pick-Place and ML10.
Gridworld was tuned separately. For gridworld, we evaluated over M, L, and XL for the standard architecture and XS, S, and M for the hypernetwork architecture.
All models solved the domain and performed similarly.
The results reported on gridworld are therefore from the smallest evaluated architecture sizes, M and XS respectively.

Additionally, on Cheetah-Dir and Walker we evaluated two-layer (rather than linear) hypernetworks, with a hidden layer of width $16$ or $32$. (Note that the input size, i.e.\ the output from the task encoder, was kept constant at 10.) In both cases, for L and XL base network sizes, adding an additional layer decreased or did not affect return.


\begin{figure}[t!]
    \centering
    \hspace{3cm}\includegraphics[height=4.5cm]{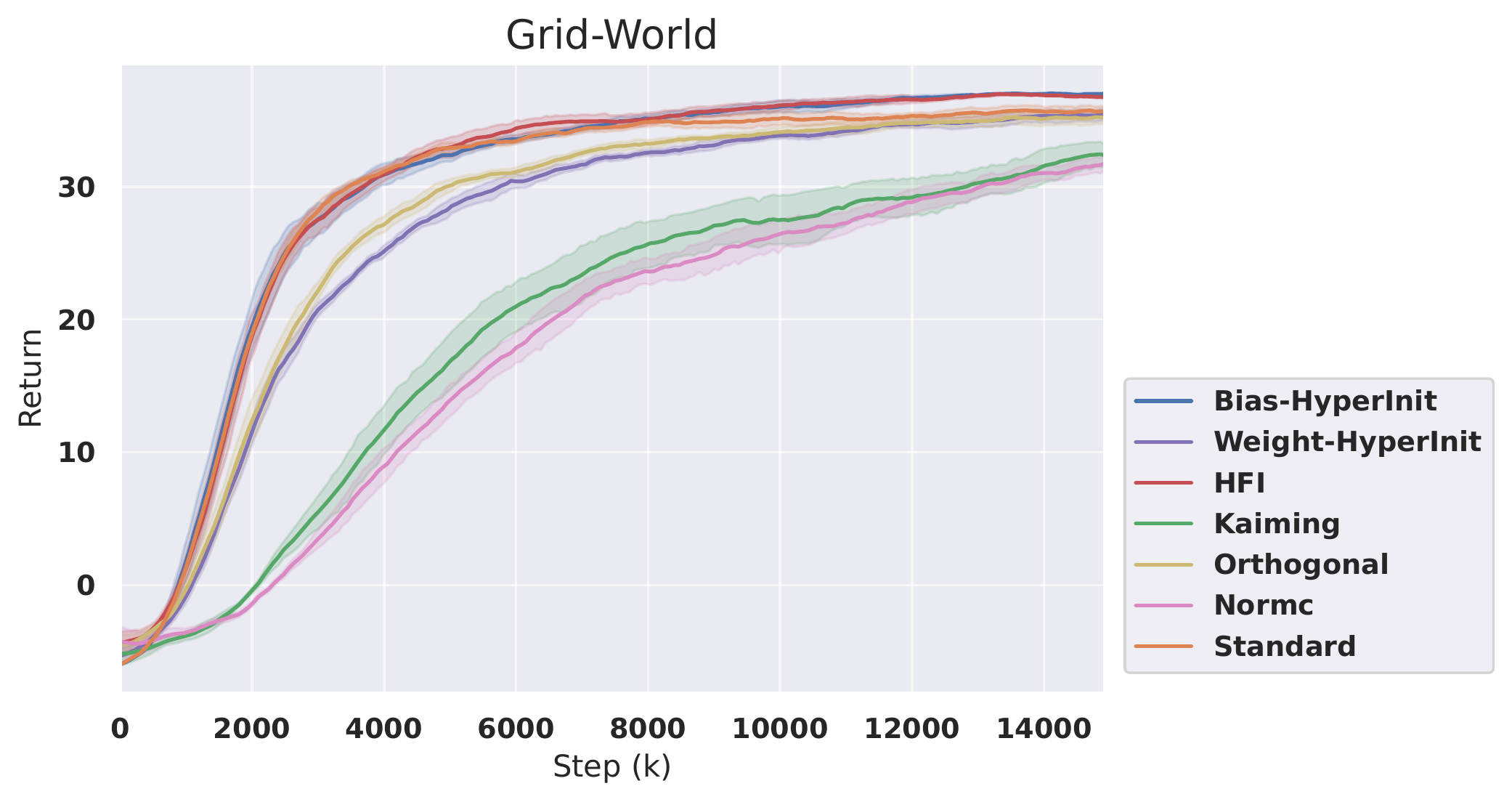} \newline
    \includegraphics[height=4.5cm]{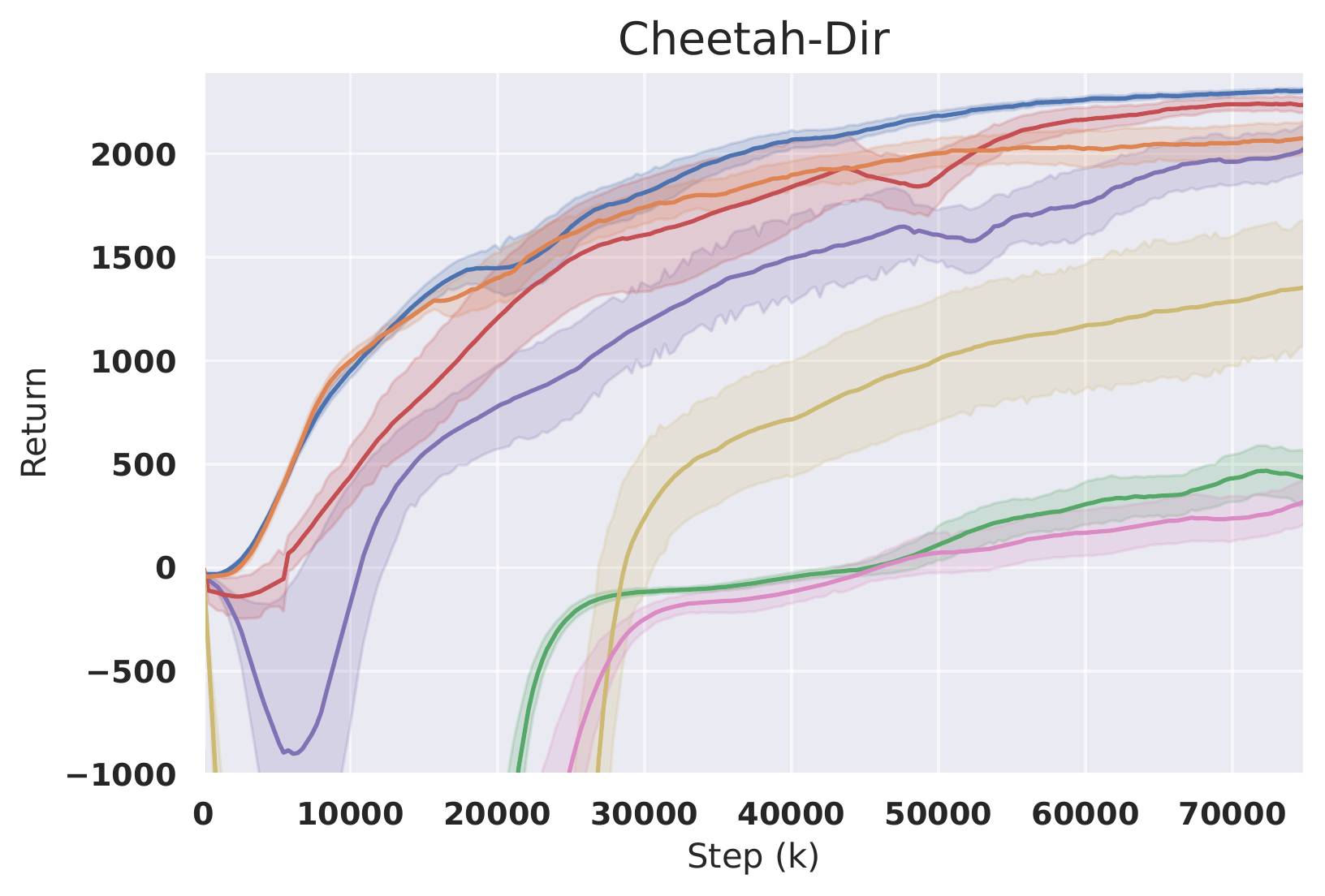}
    \includegraphics[height=4.5cm]{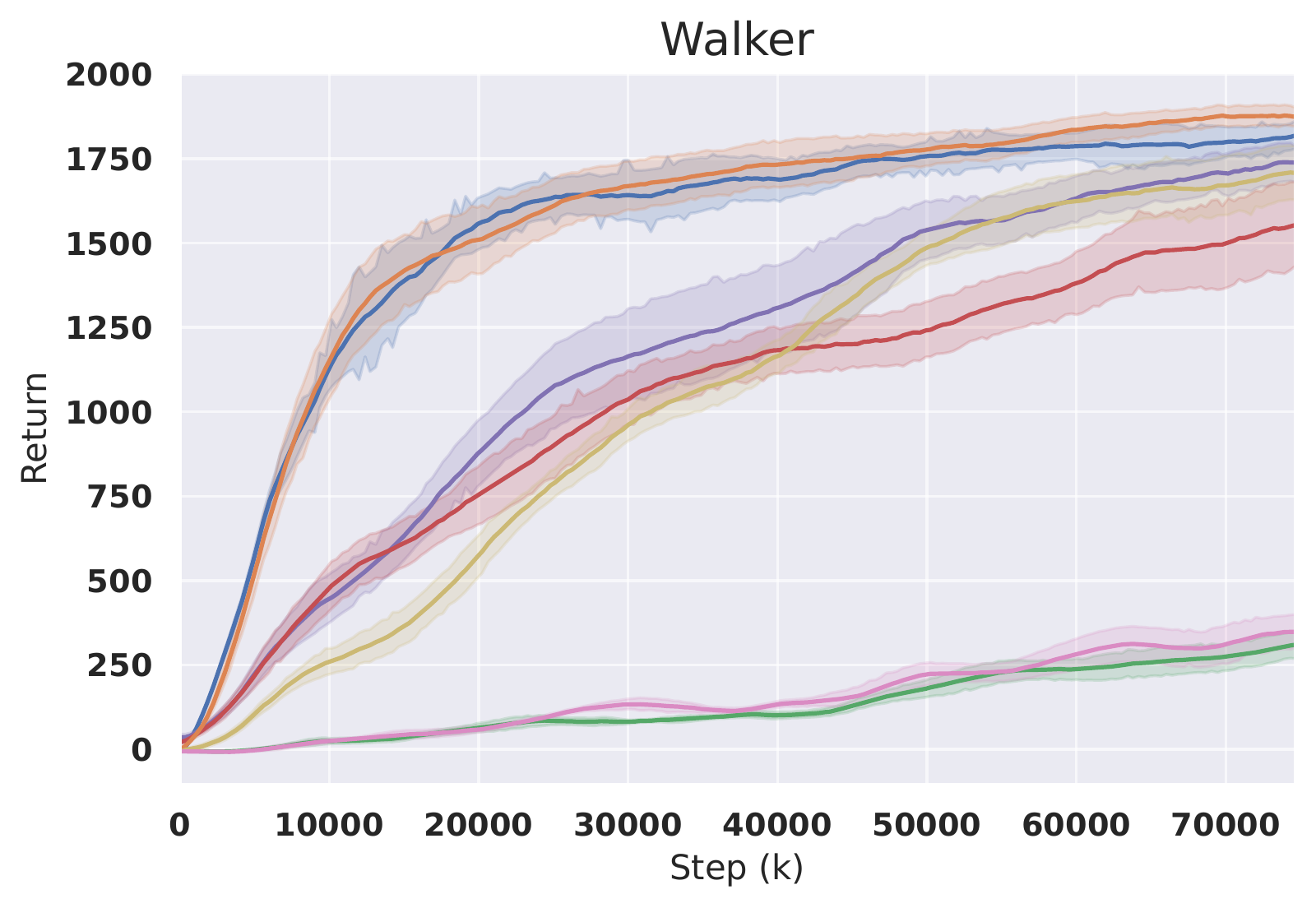} \newline
    \includegraphics[height=4.5cm]{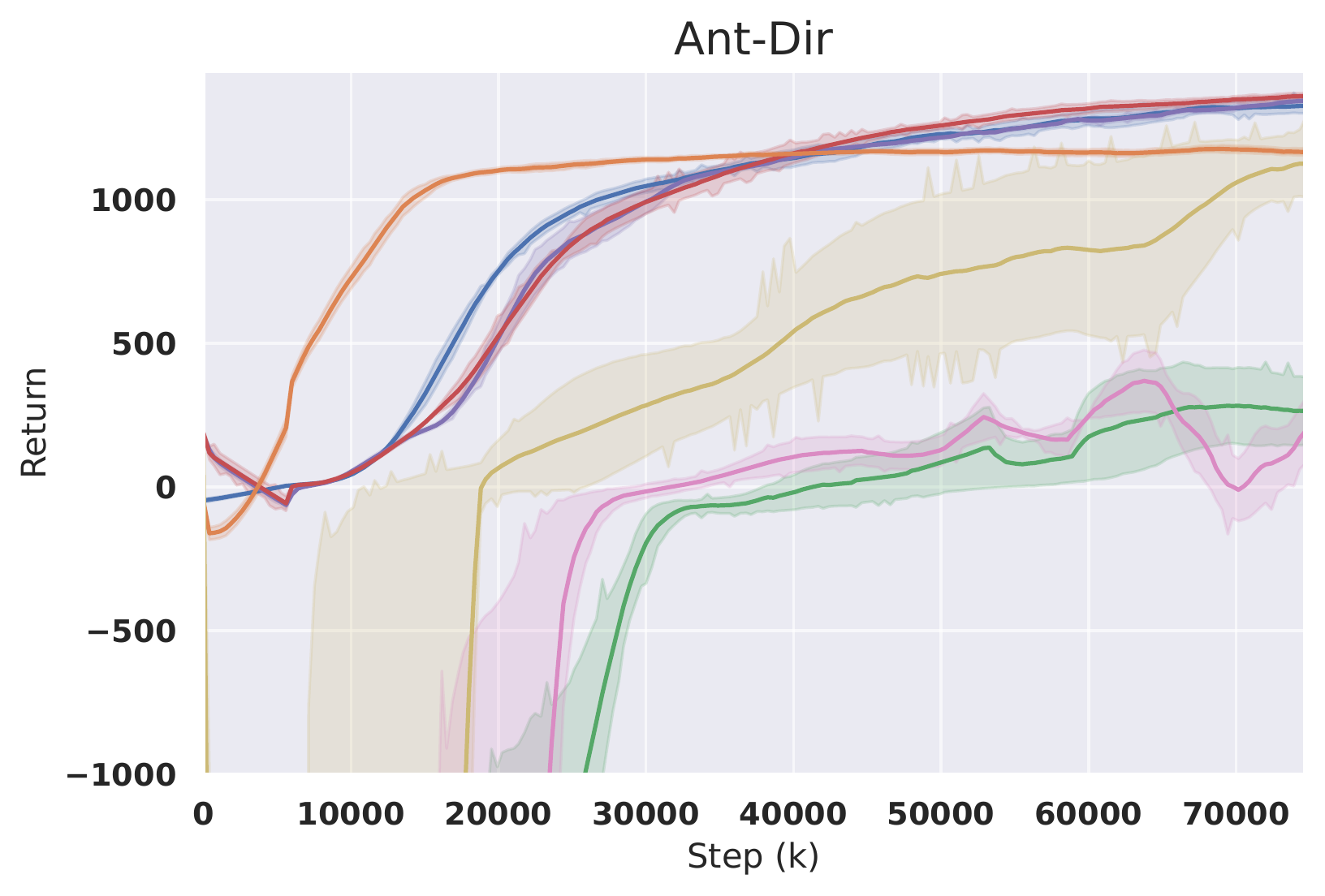}
    \includegraphics[height=4.5cm]{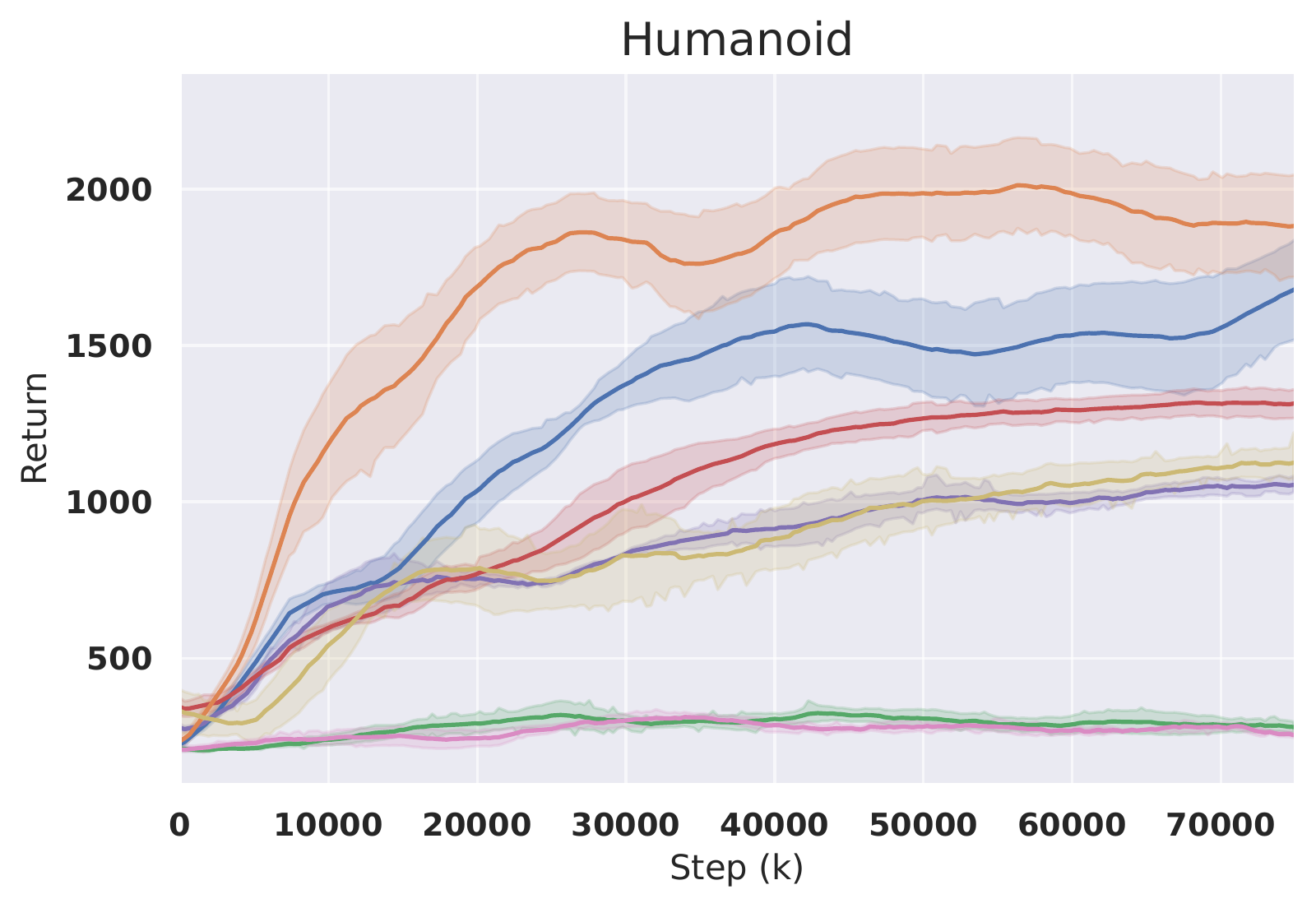}
    \newline \phantom{,} \includegraphics[height=4.52cm]{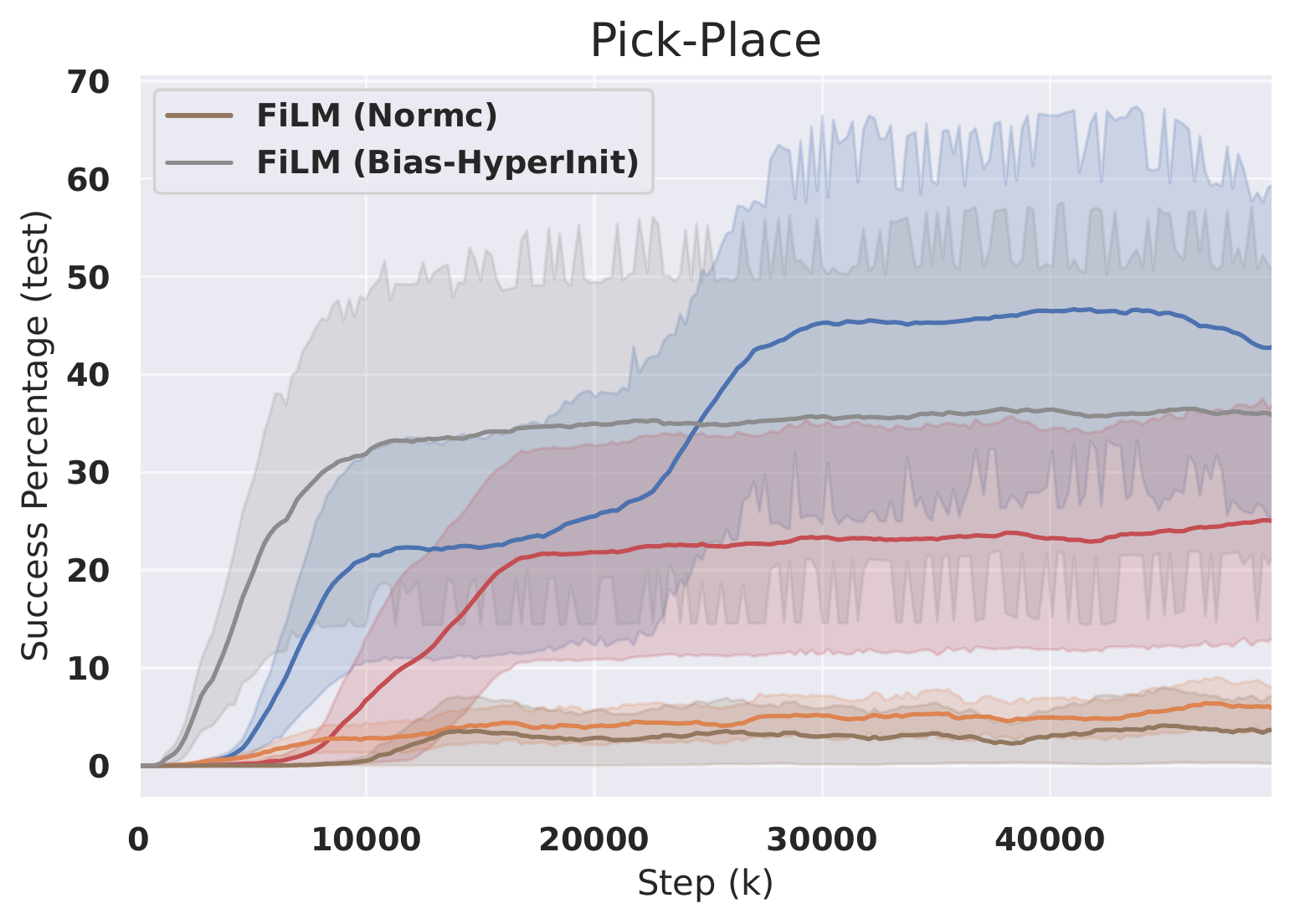} 
    \includegraphics[height=4.52cm]{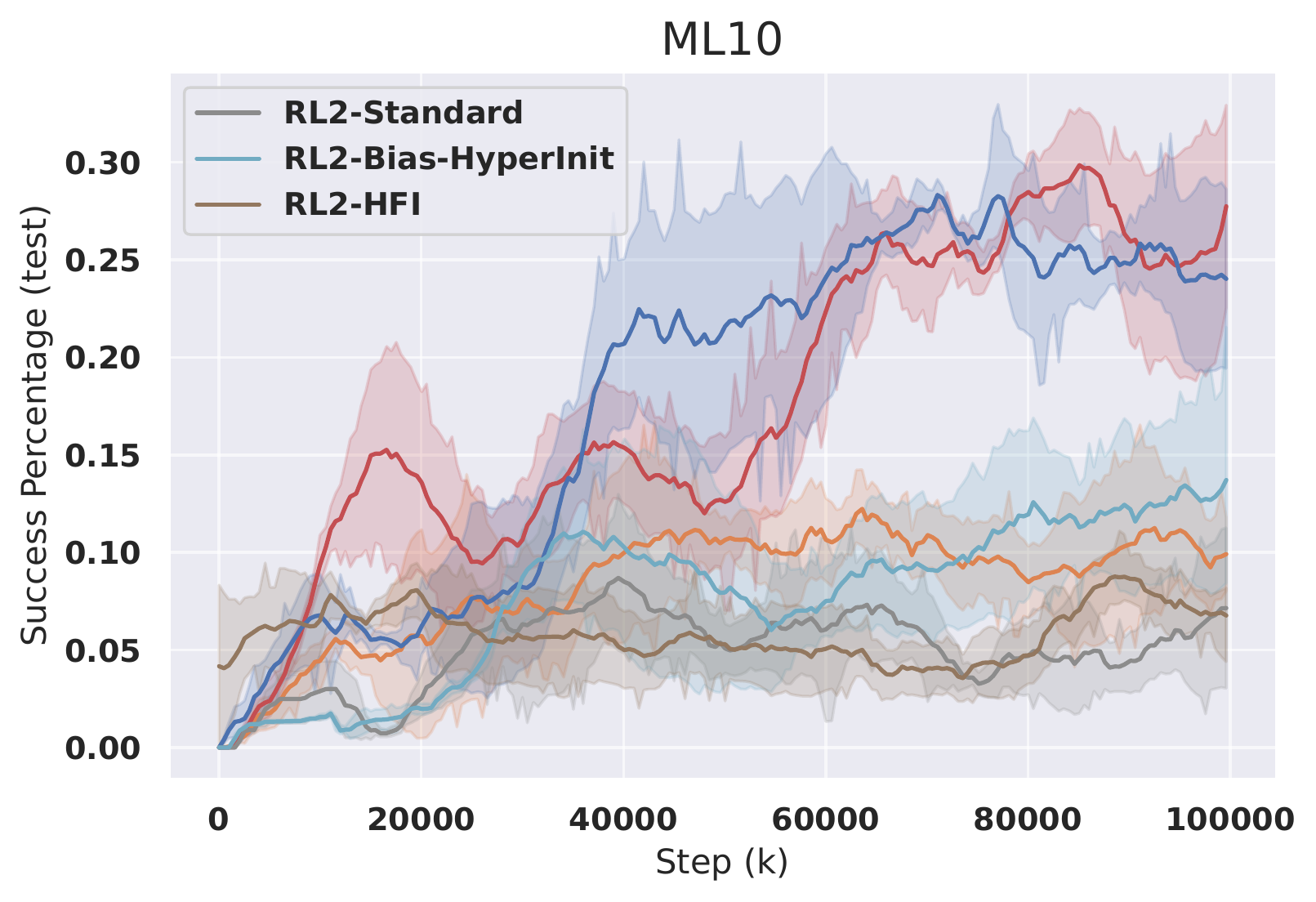}
    \caption{Learning curves for all remaining domains. (68\% confidence interval for all plots.)}
    \label{fig:all_lr_curves}
\end{figure}

\section{Learning Curves}
\autoref{fig:all_lr_curves} presents learning curves for all remaining domains and models. These results are summarized in the main paper, but are reported here for completeness. Note that reported metrics are computed over the last 1\% of data for all curves, other than ML10, which uses the last 10\% of data.




\newpage

\section{Meta-World Training Performance}

The Pick-Place and ML10 benchmarks have distinct train and test tasks. While we present and discuss test performance in the main report, \autoref{tab:train} details the train performance of the standard architecture and hypernetwork on these benchmarks.
For Pick-Place, we observe similar train and test performance.
However, we found that the ML10 train performance differed from test performance. 

\newcolumntype{R}{>{\raggedleft}p{2.5em}}
\begin{table}[ht]
    \centering
    \caption{Meta-train success percentage on Meta-World benchmarks.}
    \begin{tabular}{@{} l*{2}{R@{\hskip 0.7mm}c@{\hskip 0.7mm}r@{}} @{}}
        \toprule
        \textbf{Method} & \multicolumn{3}{r @{}}{\textbf{Pick-Place}} & \multicolumn{3}{r @{}}{\textbf{ML10}} \\
        \midrule
        Standard & $4.4$&$\pm$ & $2.1$ & $33.8$&$\pm$ & $3.0$ \\
        \cmidrule{2-7}
        HFI & $\mathbf{24.3}$&$\pm$ & $\mathbf{14.9}$ & $ 36.0$ & $\pm$ & $2.3$ \\
        Bias-HyperInit & $\mathbf{45.3}$&$\pm$ & $\mathbf{17.3}$ & $32.0$ & $\pm$ & $0.4$ \\
        \bottomrule
    \end{tabular}
    \label{tab:train}
\end{table}

At meta-train time, the standard architecture and hypernetwork with Bias-HyperInit achieve nearly identical success rates on ML10.
However, hypernetworks with Bias-HyperInit and HFI achieve significantly greater success rates on test tasks (see main report), suggesting that hyernetworks have improved generalization compared to standard models, which overfit to the train tasks.


\section{Application to FiLM} When applied to FiLM \citep{film}, the Bias-HyperInit method becomes:
\begin{enumerate}
    \item Sample parameters $\phi \sim f$, set $W=0$, and set $b$ to produce the bias parameters in $\phi$ (as in Bias-HyperInit),
    \item Set the weights of base network to those in $\phi$ (as required by FiLM),
    \item Set the remaining elements of $b$ to $1$, thereby leaving the weight distribution unchanged.
\end{enumerate}

Note that Weight-HyperInit can also be adapted such that $W$ produces distinct bias parameters for each task and scalars of 1 for all tasks, but we do not test this since
Bias-HyperInit outperforms Weight-HyperInit.

\newpage
\bibliography{refs} 
\clearpage